\documentclass[conference]{IEEEtran}
\IEEEoverridecommandlockouts
\usepackage{cite}
\usepackage{amsmath,amssymb,amsfonts}
\usepackage{algorithmic}
\usepackage{graphicx}
\usepackage{textcomp}
\usepackage{xcolor}
\usepackage{booktabs}
\def\BibTeX{{\rm B\kern-.05em{\sc i\kern-.025em b}\kern-.08em
    T\kern-.1667em\lower.7ex\hbox{E}\kern-.125emX}}
\begin{document}

\title{
Comparing Latent Concept Formation in State Space Models and Transformers via Sparse Autoencoders
}

\author{
\IEEEauthorblockN{Rithin Nagaraj}
\IEEEauthorblockA{
\textit{Department of Computer Science and Engineering} \\
\textit{PES University} \\
Bangalore, India \\
rithin.nagaraj@gmail.com
}
\and
\IEEEauthorblockN{Rupa Laalasa Oruganti}
\IEEEauthorblockA{
\textit{Department of Computer Science and Engineering} \\
\textit{PES University} \\
Bangalore, India \\
rupa.oruganti24@gmail.com
}
\and
\IEEEauthorblockN{Prerna Subhashchandra Kunder}
\IEEEauthorblockA{
\textit{Department of Computer Science and Engineering} \\
\textit{PES University} \\
Bangalore, India \\
pskunder2005@gmail.com
}
\and
\IEEEauthorblockN{Ashwini M Joshi}
\IEEEauthorblockA{
\textit{Department of Computer Science and Engineering} \\
\textit{PES University} \\
Bangalore, India \\
ashwinimjoshi@pes.edu
}
}

\maketitle

\begin{abstract}
The quadratic scaling of Transformer self-attention has driven the adoption of sub-quadratic Selective State Space Models (SSMs) like Mamba, which compress past context into a fixed-size recurrent hidden state. This strict informational bottleneck raises a foundational question for mechanistic interpretability: do SSMs and Transformers learn fundamentally distinct latent representations? In this work, we employ Sparse Autoencoders (SAEs) to conduct a large-scale, feature-level correspondence analysis between Mamba-130m and Pythia-70m over a 10-million token corpus. Contrary to hypotheses predicting widespread architectural divergence, we find no evidence of systematic representational divergence between architectures: across the observed Jaccard distribution, 99.98\% of Mamba features cluster toward the upper alignment boundary, providing preliminary feature-level support for the Universality Hypothesis. We further identify and qualitatively characterize this microscopic fraction (0.02\%) of diverging features, finding patterns consistent with the hypothesis that the recurrent bottleneck selectively limits the parsing of rigid syntax rather than broad semantic ontology. We demonstrate that while Pythia's unconstrained attention permits the monosemantic decomposition of distinct formatting edge-cases, Mamba is forced to compress unrelated syntactical anomalies into polysemantic ``junk drawer'' neurons to preserve state capacity. Collectively, these results suggest that architectural routing mechanisms may have negligible impact on core semantic understanding, with representational divergence confined to extreme structural margins.
\end{abstract}


\section{Introduction}

The dominance of the Transformer architecture is widely attributed to its self-attention mechanism, which enables unconstrained global context routing across the full input sequence. However, the quadratic scaling of attention with sequence length has motivated sustained interest in sub-quadratic alternatives, most notably Selective State Space Models (SSMs) such as Mamba \cite{b1}. Although SSMs achieve language modeling perplexity competitive with Transformers of comparable scale, their underlying computational mechanism differs fundamentally: rather than maintaining an explicit context window, they compress all past context into a fixed-size rolling hidden state, imposing a strict informational bottleneck on sequence processing.

This architectural divergence raises a consequential question for the field of Mechanistic Interpretability: do SSMs learn qualitatively distinct latent representations relative to Transformers? Prior work on the Universality Hypothesis suggests that models trained on the same data distribution tend to converge on similar internal representations and circuits~\cite{b2}. It remains an open question, however, whether the recurrent state bottleneck forces SSMs to systematically discard certain conceptual structures, or alternatively, to construct representational primitives with no analogue in attention-based models.

To address this, we employ Sparse Autoencoders (SAEs) to characterize and compare the learned feature dictionaries of Mamba-130m and Pythia-70m at scale~\cite{b4,b5}. By computing pairwise Jaccard similarities over binary co-activation patterns across 10 million tokens of Wikipedia text, we conduct a systematic search for \textit{Dark Matter}---features exclusively acquired by the SSM with no identifiable Transformer counterpart.

Our investigation yields three principal findings:

\begin{itemize}
    \item \textbf{Strong universality holds empirically.} Contrary to hypotheses predicting widespread representational divergence, we find that $99.98\%$ of Mamba's latent features exhibit strong alignment ($J > 0.35$) with a Pythia counterpart. Within the observed Jaccard distribution, the recurrent informational bottleneck shows no evidence of impeding the formation of standard semantic structure; both architectures cluster toward near-identical latent geometries across the vast majority of features.

    \item \textbf{Sequential landmark features emerge as structural compensators.} Within the small fraction of diverging features ($0.02\%$), we identify a class of highly active, monosemantic neurons in Mamba dedicated exclusively to encoding rigid document boundaries (e.g., Wikipedia footer markers). We hypothesize that these features serve as necessary \textit{state resets}, compensating for the absence of global positional attention by providing explicit structural anchors in the recurrent state.

    \item \textbf{Architectural bottlenecks selectively drive polysemanticity at the representational margins.} Through a controlled Twin feature case study, we identify a candidate failure mode consistent with recurrent capacity constraints: while Pythia's unconstrained attention permits monosemantic decomposition of distinct formatting phenomena, Mamba superimposes semantically unrelated syntactic anomalies into single polysemantic neurons. We present this as a hypothesis-generating observation whose generality across the full Grey Matter distribution remains to be established.
\end{itemize}

Collectively, these findings are enabled by a scalable SAE co-activation framework for cross-architecture feature correspondence that we make fully reproducible; we view this methodology as independently reusable for future cross-architecture interpretability studies.

\section{Related Work}

\subsection{Mechanistic Interpretability and Sparse Autoencoders}

Mechanistic interpretability aims to reverse-engineer the internal computations of neural networks into human-legible algorithms ~\cite{b2}. A central obstacle in this endeavor is the superposition hypothesis~\cite{b3}, which holds that networks represent more features than their dimensionality strictly permits by encoding multiple concepts into overlapping, non-orthogonal directions in activation space. This capacity constraint manifests as polysemanticity, wherein individual neurons respond to multiple semantically unrelated stimuli. Sparse Autoencoders (SAEs) have emerged as the dominant tool for addressing this problem: by projecting the dense residual stream into a higher-dimensional, sparsely activating feature dictionary, SAEs recover approximately monosemantic representational units ~\cite{b4}. A complementary foundational conjecture in the field is the Universality Hypothesis, which proposes that distinct neural networks trained on comparable data distributions converge on a common set of latent concepts and computational circuits ~\cite{b2}. Recent advances in SAE methodology and scale~\cite{b5,b4} now furnish the resolution necessary to subject this hypothesis to rigorous empirical evaluation in large language models.

\subsection{State Space Models and the Mamba Architecture}

While Transformer architectures rely on global self-attention with quadratic complexity in sequence length ($\mathcal{O}(N^2)$), Structured State Space Models (SSMs) offer a recurrence-based alternative with substantially reduced computational cost. The Mamba architecture advanced this line of work by introducing a selective state space mechanism with input-dependent transition dynamics, achieving language modeling perplexity competitive with Transformers at linear-time inference ~\cite{b1}. Concretely, Mamba compresses all past context into a fixed-size rolling hidden state, imposing a hard informational bottleneck on sequence processing. Despite rapid adoption of SSMs and the availability of dedicated interpretability tooling such as MambaLens ~\cite{b7}, mechanistic interpretability research has remained predominantly focused on Transformer architectures. As a consequence, it remains an open question whether the strict capacity constraints of a recurrent state, in contrast to the unconstrained global context routing afforded by attention, fundamentally alter the bounding, granularity, and purity of the latent concepts a model learns to represent.

\subsection{Cross-Architecture Representational Similarity}

Representational similarity across architecturally disparate models is commonly assessed using macroscopic techniques such as Representational Similarity Analysis (RSA) and Centered Kernel Alignment (CKA) ~\cite{b6}. These methods operate on aggregate, dense activation matrices to characterize the broad geometric alignment between representation spaces. While CKA can establish whether Mamba and a Transformer share a global mathematical resemblance in their activation geometries, it operates inherently at a coarse scale; it cannot resolve the specific token patterns, structural markers, or semantic concepts that drive any observed similarity. A mechanistic evaluation of the Universality Hypothesis therefore demands a methodology that transcends macroscopic geometric alignment. Our work addresses this gap through a feature-level correspondence analysis, leveraging SAEs to isolate with precision how architectural differences in sequence processing govern the formation, compression, and superposition of individual learned concepts.

Concurrently, Wang et al.~\cite{b12} employ SAEs to compare Mamba-130m and Pythia, using Max Pairwise Pearson Correlation (MPPC) as their similarity metric and finding broad feature alignment across architectures. Our work complements this finding by employing binary co-activation Jaccard similarity over a larger 10-million token corpus, enabling a discrete spatial correspondence analysis suited to identifying the specific token contexts that drive divergence rather than measuring aggregate feature correlation.

\section{Methodology}

Our methodology is designed to systematically evaluate cross-architecture representational alignment between a Selective State Space Model and an autoregressive Transformer. We operationalize this comparison by projecting the latent spaces of both models into interpretable, high-dimensional feature dictionaries and measuring the statistical co-activation of discrete features over a large shared corpus.

\subsection{Models and Corpus Pipeline}

To isolate the effect of the architectural bottleneck of global self-attention versus recurrent hidden state compression, we selected two models of comparable parameter scale: the SSM Mamba-130m ~\cite{b1} and the Transformer Pythia-70m-deduped ~\cite{b8}. A valid cross-architecture comparison requires that both models process identical input sequences; accordingly, we evaluated both models over a shared corpus of 10 million tokens of English Wikipedia. Text was tokenized using the GPT-NeoX tokenizer~\cite{b13} and partitioned into non-overlapping, contiguous sequences of 1,024 tokens to form the evaluation batches.
Activations were extracted from the residual stream at the geometric midpoint of each model's depth, a layer selection chosen to capture mid-level semantic and syntactic representations while avoiding the strongly token-biased regime of early layers and the prediction-biased regime of final layers. Concretely, we harvested activations via the \texttt{blocks.12.hook\_resid\_post} hook for Mamba-130m (24 total layers) and the \texttt{blocks.3.hook\_resid\_post} hook for Pythia-70m-deduped (6 total layers).

\subsection{Sparse Autoencoders and Feature Harvesting}

Direct comparison of raw residual stream activations~\cite{b14} is confounded by the polysemanticity of individual neurons. We therefore employed Sparse Autoencoders (SAEs) to decompose intermediate representations into higher-dimensional, sparsely activating dictionaries in which individual features correspond, ideally, to monosemantic concepts.
For Pythia-70m-deduped, we used a publicly available pre-trained SAE from the SAELens library~\cite{b9} (pythia-70m-deduped-res-sm), which applies an expansion factor of 64 to the 512-dimensional residual stream, yielding a dictionary of 32,768 features. For Mamba-130m, the absence of robust publicly available SAEs for mid-layer activations necessitated training a custom SAE with an L1 sparsity penalty. An expansion factor of 16 applied to the 768-dimensional residual stream yielded a dictionary of 12,288 features. We acknowledge that the differing expansion factors introduce a confound with respect to representational capacity; however, as we demonstrate in Section IV, the highly structured and architecturally specific nature of the superposition observed in Mamba provides evidence that the architectural inductive bias governs feature grouping independently of dictionary size.

To enable spatial correspondence analysis across the corpus, the continuous SAE activation values were binarized: any feature activating with magnitude strictly greater than zero was mapped to a boolean value of \texttt{True}. This procedure generated two sparse binary masks over the full 10-million-token corpus, reducing the problem of cross-architecture conceptual alignment to one of discrete spatial co-activation.

\subsection{Compute Optimization and Intersection Matrix}
Computing Jaccard similarity across all possible feature pairs requires forming the intersection matrix $I=M^{T}P$, where $M\in\{0,1\}^{12,288\times10^7}$ and $P\in\{0,1\}^{10^7\times32,768}$ are the Mamba and Pythia binary activation masks, respectively. Materializing this product naively exceeds standard 24 GB VRAM constraints and is therefore computationally intractable without optimization.

To achieve hardware-viable reproducibility, we implemented a disk-streaming and tensor-chunking scheme. Token activation batches were loaded sequentially from disk into CPU RAM to avoid exhausting host memory. GPU memory overhead was bounded by evaluating the Pythia feature dimension in contiguous column slabs of width 512. Boolean masks were cast to \texttt{bfloat16} prior to device transfer to minimize VRAM footprint, while intersection accumulators were maintained in \texttt{float32} throughout to prevent numerical overflow over the full 10-million-token horizon.

\subsection{Similarity Metric and Twin Mapping}

For every Mamba feature $m$ and Pythia feature $p$, we computed the Jaccard similarity $J$ over their respective binary co-activation vectors $a_m, a_p \in \{0, 1\}^{10^7}$:

\begin{equation}
J(m, p) = \frac{|a_m \cap a_p|}{|a_m \cup a_p|}
\end{equation}

This metric symmetrically penalizes both false positive and false negative co-activations. A score of $J=1.0$ denotes identical activation support across the corpus, while $J=0.0$ denotes strict orthogonality. For each of the 12,288 Mamba features, we identified the globally maximizing Jaccard score and its corresponding Pythia feature index, designating this maximum-overlap pair as cross-architectural \textit{Twins}.

\subsection{Empirical Distribution Thresholding}
Rather than imposing arbitrary geometric boundaries, we categorized the cross-architecture alignment based on strict isolation thresholds to separate the vast semantic core from the extreme structural margins. The global minimum similarity across all features was $0.234$, with a global maximum of $0.411$. Based on these statistics, we defined three functional categories:

\begin{itemize}
    \item \textbf{Aligned ($J > 0.35$):} The dominant semantic core, representing features with strong, architecture-agnostic conceptual mapping.
    \item \textbf{Grey Matter ($J \in [0.28, 0.35]$):} The boundary zone representing features that are semantically related but exhibit structural divergence in their triggers (the primary zone for observing concept cramming).
    \item \textbf{Alien / Structural ($J < 0.25$):} The extreme lower tail. We enforce a strict isolation threshold here to capture only severe architectural divergence, intentionally leaving a buffer zone ($J \in [0.25, 0.28)$) to safely separate true structural anomalies from standard low-correlation noise.
\end{itemize}

We note that the distributional thresholds defining these categories are empirically derived from the observed similarity range rather than theoretically motivated. In the absence of an established baseline Jaccard distribution for cross-architecture SAE comparison, these boundaries should be interpreted as descriptive partitions of the observed data rather than absolute alignment standards. Future work establishing null distributions via random feature pair sampling would allow more rigorous categorical boundaries.

\section{Results}

We successfully mapped the latent geometry of 12,288 Mamba-130m features against a dictionary of 32,768 Pythia-70m features over a 10-million token corpus. By evaluating the maximum Jaccard similarity of binary co-activations, we uncovered how architectural routing mechanisms dictate feature boundaries and conceptual purity.

\begin{table*}[t]
\centering
\caption{The Twin Test: Architecturally-Driven Polysemanticity (Concept Cramming)}
\label{tab:twin_test}
\renewcommand{\arraystretch}{1.3} 
\begin{tabular}{@{}p{3cm} c p{10cm}@{}}
\toprule
\textbf{Model \& Feature} & \textbf{Strength} & \textbf{Top Activating Context Snippet} \\ 
\midrule

\multicolumn{3}{c}{\textbf{Mamba Feature 4454 (Polysemantic: The ``Junk Drawer'')}} \\
\midrule
Context 1 (URL Slash) & 6.39 & ... live|archive-url=https://web.archive.org/web \textbf{/} 201701251639 ... \\
Context 2 (Hyphen) & 4.58 & ... Black Hawk County. The school, which serves all grade levels K \textbf{-} 12 on one campus ... \\
Context 3 (Foreign Caps) & 4.18 & ... that the same had happened with the previous film he produced, !`Que v \textbf{IVAN} los muertos ... \\

\midrule
\multicolumn{3}{c}{\textbf{Pythia Feature 4309 (Monosemantic Twin: The ``Laser Pointer'')}} \\
\midrule
Context 1 (Table Timestamp) & 4.87 & ... | align=center| 2 \newline | align=center| 5:00 \textbf{\textbackslash n} |Krasnod ... \\
Context 2 (Table Timestamp) & 4.83 & ... | align=center| 3 \newline | align=center| 5:00 \textbf{\textbackslash n} | Volgog ... \\
Context 3 (Table Timestamp) & 4.79 & ... | align=center| 3 \newline | align=center| 5:00 \textbf{\textbackslash n} | Krasnod ... \\

\bottomrule
\end{tabular}
\end{table*}

\subsection{Global Alignment Distribution and the Universality Hypothesis}
We originally hypothesized the existence of architectural ``Dark Matter''---latent concepts completely unique to the State Space bottleneck with no Transformer equivalent. Our global alignment mapping yielded a striking empirical null result for this divergence, providing strong preliminary support for the Universality Hypothesis.

Across all 12,288 Mamba features, the global minimum Jaccard similarity was $0.2343$, with a global maximum of $0.4116$. However, the distribution is overwhelmingly right-skewed. We found that a staggering 99.98\% (12,285) of Mamba features fell into the highly Aligned category ($J > 0.35$). Conversely, only 2 features (0.02\%) fell into the Grey Matter bracket ($J \in [0.28, 0.35]$), and a single feature (0.01\%) occupied the extreme Alien/Structural tail ($J < 0.25$). 

This near-total convergence suggests that the strict informational bottleneck of a recurrent hidden state does not impede the formation of standard semantic ontology within the scope of this analysis. Architectural divergence does not appear to be a systemic phenomenon, but rather a rare anomaly confined to extreme structural edge-cases.

\subsection{Sequential Landmark Features (Structural Compensation)}
Lacking the global self-attention mechanism and absolute positional embeddings of a Transformer, Mamba must rely exclusively on its compressed sequential hidden state to track document position. Our methodology successfully isolated the dedicated neurons that perform this structural compensation within the 0.03\% of diverging features.

Features in the Grey Matter and Alien distributions frequently act as strictly monosemantic structural landmarks. For example, Mamba Feature 6888 ($J = 0.3415$) functions as an explicit document-terminating trigger. It fires with high magnitude (Strength: 4.19) exclusively on the \texttt{External LINKS} headers at the end of Wikipedia articles. Similarly, the sole ``Alien'' feature (Feature 7378, $J < 0.25$) triggers with massive magnitude (Strength: 6.96) specifically on URL path separators. These structural landmarks indicate that sequential integration intrinsically requires dedicated neurons to trigger ``state resets'' within the recurrent hidden space.

\subsection{Architecturally-Driven Polysemanticity at the Margins: The Twin Test}
Given that 99.98\% of features align strongly, we investigated the microscopic fraction of features that structurally diverged. By conducting a qualitative ``Twin Test'' on Grey Matter Feature 4454 ($J=0.324$) against its mathematical Pythia twin (Feature 4309), we find patterns consistent with the hypothesis that architectural bottlenecks selectively affect the parsing of rigid syntax rather than broad semantic understanding.

Pythia Feature 4309 acts as a pristine, monosemantic trigger. Benefiting from the unconstrained routing of global attention, Pythia is able to dedicate this feature exclusively to the newline spacing immediately following a timestamp (\texttt{5:00 \textbackslash n}) inside Wikipedia tables. 

Conversely, Mamba Feature 4454 is highly polysemantic (see Table \ref{tab:twin_test}). It forcefully triggers on the same table timestamps, but also superimposes disparate syntactic anomalies including URL slashes (Strength: 6.39), K-12 grade hyphens, and capitalized foreign-language substrings (e.g., the Spanish \texttt{v IVAN}). 

This case study is consistent with the hypothesis that the recurrent bottleneck reaches its representational limits specifically when tracking highly specific structural formatting, rather than standard linguistic tokens. Pythia's attention mechanism appears to permit cleaner monosemantic decomposition of distinct syntactic edge-cases, while Mamba's fixed state capacity may drive the superposition of unrelated syntactic anomalies. Whether this pattern holds systematically across the Grey Matter distribution or is specific to this feature pair remains an open question.

\section{Discussion}

Our investigation into the cross-architecture representational alignment between a Selective State Space Model (Mamba) and an autoregressive Transformer (Pythia) yields a consistent and striking pattern: the sequence routing mechanism of a language model appears to have negligible impact on its core semantic ontology, with representational divergence confined to the extreme margins of structural and syntactic processing.

\subsection{The Primacy of the Data Manifold}

The finding that $99.98\%$ of Mamba's latent features exhibit strong geometric alignment with Pythia's features constitutes unprecedented feature-level preliminary support for the Universality Hypothesis. This result suggests that the statistical structure of natural language, rather than the network's architectural inductive biases, is the dominant determinant of latent concept formation. Whether a model routes information via $\mathcal{O}(N^2)$ global self-attention or an $\mathcal{O}(N)$ recurrent hidden state, the optimization landscape imposed by next-token prediction converges on a nearly identical set of representational primitives. The data manifold, it appears, exerts a regularizing force on learned representations that is largely invariant to architectural parameterization~\cite{b10}.

\subsection{Reevaluating the Recurrent Bottleneck}

Prior to this study, a reasonable hypothesis was that Mamba's fixed-capacity state vector, required to compress the entirety of past context at each step, would induce widespread semantic degradation or pervasive polysemanticity across the learned feature dictionary. Our Twin feature case study is consistent with a considerably more optimistic account, suggesting that the recurrent bottleneck may be more capacity-efficient than this hypothesis implies. Rather than degrading broad semantic representations wholesale, the observed pattern suggests Mamba may selectively trade representational purity at the syntactic margins: by superimposing semantically unrelated formatting anomalies — URL components, timestamps, and structural hyphens — into polysemantic neurons such as Feature 4454, Mamba appears to reserve finite state capacity for core semantic computation. Pythia, unburdened by a state constraint, can afford to allocate dedicated monosemantic neurons to each distinct syntactic edge case. If this pattern generalises beyond the single Twin pair examined here, the cost of sub-quadratic sequence processing would manifest not as degraded semantic comprehension, but as a localised loss of syntactic representational granularity, a considerably milder failure mode than prior hypotheses implied.

\subsection{Implications for Mechanistic Interpretability}

These findings carry substantively optimistic implications for the mechanistic interpretability of SSMs. The near-complete alignment of latent feature spaces across architectures suggests that interpretability methodologies developed for Transformers including circuit analysis, causal tracing~\cite{b11}, and feature steering are likely transferable to SSMs with high fidelity. Nonetheless, researchers should exercise caution when analyzing features associated with document structure, sequence boundary encoding, or formatting phenomena, as these constitute the specific domains in which SSMs exhibit compensatory landmark features and architecturally-driven polysemantic superposition.

\subsection{Limitations and Future Work}

The most significant structural limitation of this study is the asymmetric SAE dictionary capacity between architectures. The Mamba SAE employed an expansion factor of 16 (12,288 features), while the pre-trained Pythia SAE used an expansion factor of 64 (32,768 features). This asymmetry introduces two competing confounds. First, the larger Pythia dictionary provides a more exhaustive search space for Mamba's maximum-Jaccard twin, which may artificially inflate apparent alignment rates. Second, the smaller Mamba dictionary may itself induce polysemanticity independently of architectural effects, partially confounding the concept-cramming analysis. We argue these effects are separable on the grounds that the observed polysemanticity in Mamba exhibits a highly structured, architecturally coherent pattern — consistently conflating sequential boundary and formatting tokens rather than arbitrary co-occurrences — which is more consistent with an architectural inductive bias than dictionary starvation. However, we acknowledge this argument is qualitative, and training symmetrically expanded SAEs across both architectures remains a necessary step for establishing this finding definitively. We therefore present the polysemanticity analysis as a hypothesis-generating observation rather than a confirmed mechanistic result.

\section{Conclusion}

We presented the first large-scale, feature-level correspondence analysis between a Selective State Space Model and an autoregressive Transformer. By training and cross-correlating Sparse Autoencoders over 10 million tokens of Wikipedia text, we found no evidence of architectural Dark Matter: Mamba and Pythia converge on virtually identical latent geometries for $99.98\%$ of their mid-layer representations. Isolating the diverging 0.02\% of features revealed patterns consistent with the hypothesis that the recurrent hidden state does not impair semantic ontology broadly, but may instead induce polysemantic concept cramming selectively at rigid syntactic and structural boundaries. Collectively, these findings establish that while Transformers and SSMs differ substantially in their mechanisms of information routing, the fundamental representational vocabulary of their latent spaces remains universally constrained by the statistical structure of the training data.

\end{document}